\documentclass[11pt,a4paper]{article}
\usepackage[hyperref]{acl2020}
\usepackage{times}
\usepackage{latexsym}

\usepackage{microtype}
\usepackage{graphicx}
\usepackage{xcolor}
\usepackage{float}
\usepackage{tabularx}
\aclfinalcopy 

\title{A comprehensive study of on-device NLP applications - VQA, automated Form filling, Smart Replies for Linguistic Codeswitching}

\author{Naman Goyal \\
  Columbia University \\
  \texttt{ng2848@columbia.edu} \\
  }

\date{}

\begin{document}
\maketitle
\begin{abstract}
Recent improvement in large language models, open doors for certain new experiences for on-device applications which were not possible before. In this work, we propose 3 such new experiences in 2 categories. First we discuss experiences which can be powered in screen understanding i.e. understanding whats on user screen  namely - (1) visual question answering, and (2) automated form filling based on previous screen. The second category of experience which can be extended are smart replies to support for multilingual speakers with code-switching. Code-switching occurs when a speaker alternates between two or more languages. To the best of our knowledge, this is first such work to propose these tasks and solutions to each of them, to bridge the gap between latest research and real world impact of the research in on-device applications.
\end{abstract}

\section{Introduction}

Recent advancements in large language models in Document AI \citep{xu2020layoutlm, li2021markuplm}, Dialogue Generation \citep{peng2022godel}, information extraction opens doors to much powerful and useful experiences on device application, much better than existing solutions. Currently the 2 main applications of NLP on a device are (1) digital assistants and (2) automated smart replies. This work gives a comprehensive idea of newer experiences which can be supported on a mobile device to ease user lives. 

The first such experience are related to on-device screen understanding. The latest research on layout based understanding of screen content and help power newer capabilities on device including VQA, smarter form filling and information sharing and better accessibility usage for visually challenged users.

The second experience, is extending the capability of smart replies systems. Smart replies refers to automatically generating shorter responses for an incoming email or an ongoing conversation which a user assists with quicker response in a large number of scenarios. The current research in this domain is similar to dialogue generation, but the current responses are much more generic and only applicable in highly formal and limited languages setting. This works discusses new applications for smart replies  (1) smart replies for users with linguistic code switching (2) personalized smart reply generation based on learning knowledge about a user from the conversation history

In linguistics, code-switching or language alternation occurs when a speaker alternates between two or more languages, or language varieties, in the context of a single conversation or situation. Multilingual (speakers of more than one language) sometimes use elements of multiple languages when conversing with each other. Thus, code-switching is the use of more than one linguistic variety in a manner consistent with the syntax and phonology of each variety.

The current challenges in the proposed space is lack of datasets and evaluation benchmarks for the newly proposed experiences including but not limited to, on-device screen understanding for form filling, smart replies for users with linguistic code switching, and personalized smart reply generation based on external grounded knowledge. Constructing such datasets requires large scale collections from user habits and special annotations for the task, and then coming up with good evaluation benchmarks for the tasks, 

This is the first work which discusses such newer experiences and then proposes solutions based on the current research. The notable contributions being - 

\begin{enumerate}
    \item Proposal of screen based understanding for Visual Question answering
    \item Proposal of screen based understanding for in-form filling tasks.
    \item Proposal of smart replies for code switching for multilingual speakers.
\end{enumerate}

\section{Screen Understanding}
\subsection{Introduction}

Screen understanding is modelled as understanding the input screen image and associated text on the screen. It is a multimodal learning task. Recently, there has been recently a lot of progress in Document AI for work related to document understanding. If we model an input app view as document, we can use the recent techniques in document AI. Document AI refers to given a input document, techniques for automatically (1) reading, (2) understanding, (3) analyzing it. It's a challenging task due to the diversity of layouts and formats, inferior quality of scanned document images complexity of template structures. For rest of the discussion we will consider an app view (the screenshot + screen text) as an document.

\subsection{Related Work}
\begin{figure}[t]
\centering
\includegraphics[width=1.0\linewidth]{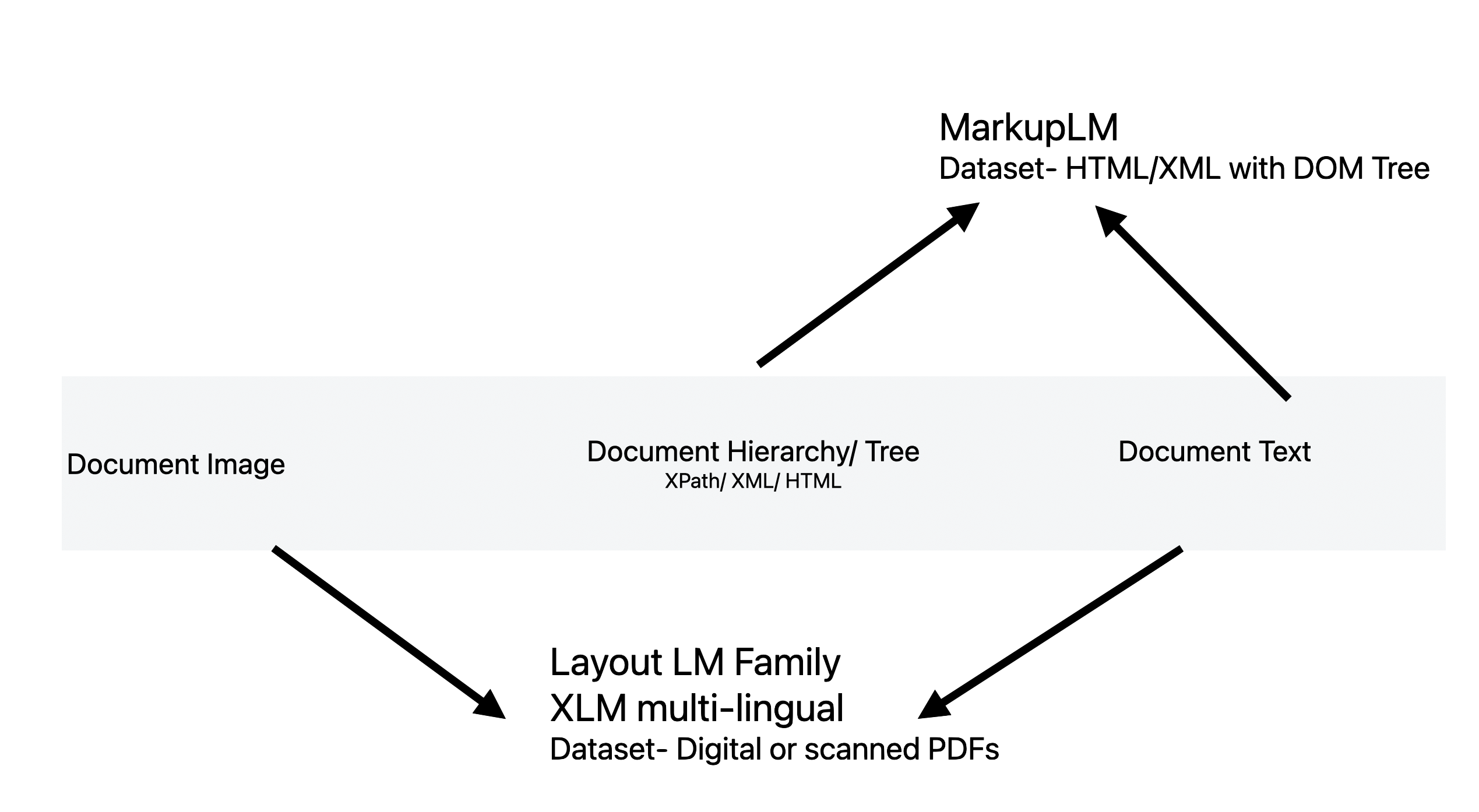}
 \caption{Families of Document AI model based on information}
 \label{fig:laymodel}
\end{figure}

There are 2 families of models proposed - LayoutLM \citep{huang2022layoutlmv3} and MarkupLM \citep{li2021markuplm} as seen in figure \ref{fig:laymodel}.
The LayoutLM family used a rendered image and the document text to extract the answer. More useful for PDFs which are rendered the same irrespective of the viewing device.
MarkupLM family uses the idea that the same HTML document could be rendered in different ways based on viewport screen size. Hence to generate they use XPath (XML Path Language) which is directly extracted from the view hierarchy.

\begin{figure*}[hb]
    \centering
    \includegraphics[width=0.9\linewidth]{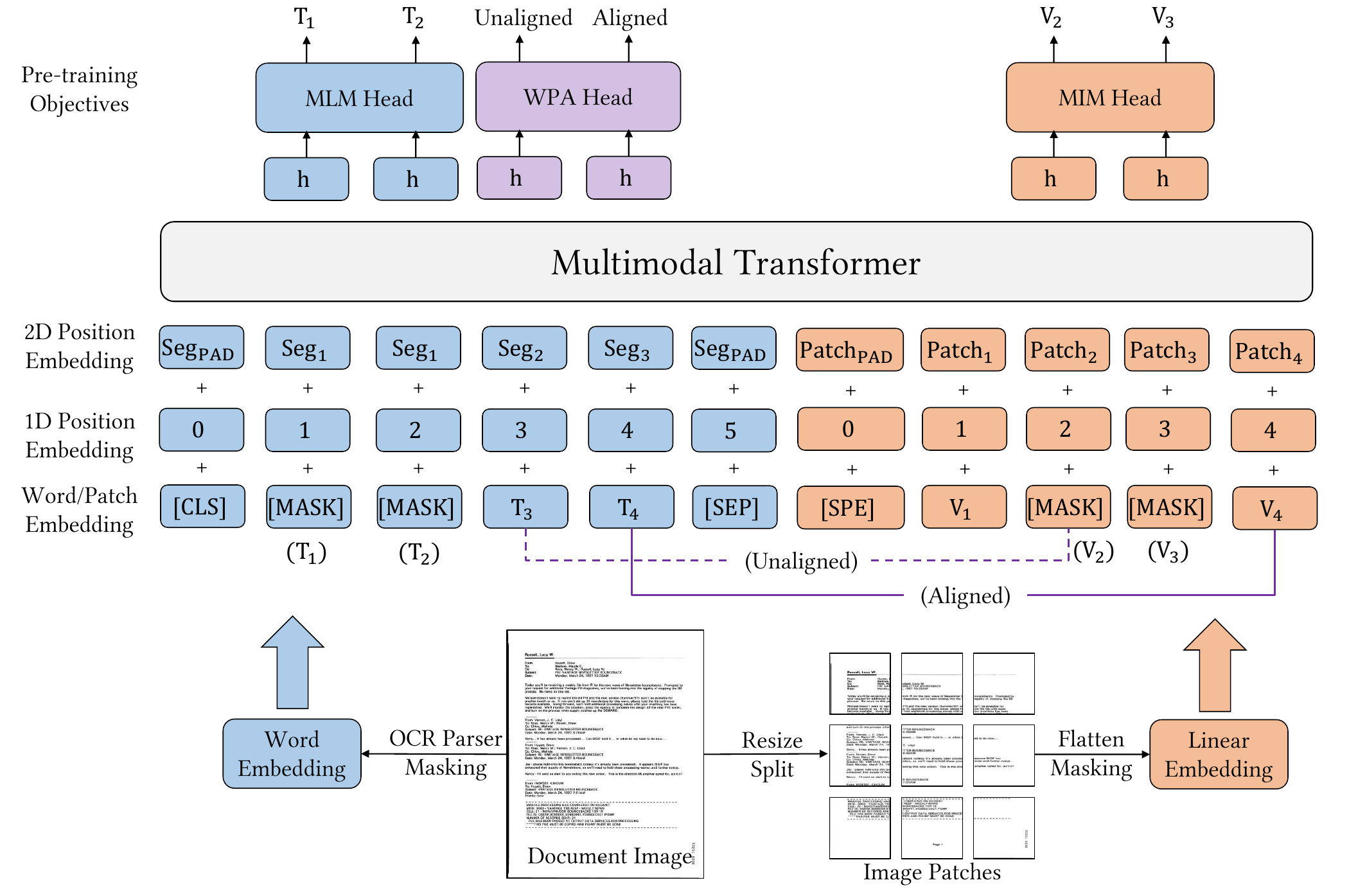}
    \caption{
    \textbf{The architecture and pre-training objectives of LayoutLMv3.}
    LayoutLMv3 is a pre-trained multimodal Transformer for Document AI with unified text and image masking objectives.
    Given an input document image and its corresponding text and layout position information, the model takes the linear projection of patches and word tokens as inputs and encodes them into contextualized vector representations.
    LayoutLMv3 is pre-trained with discrete token reconstructive objectives of Masked Language Modeling (MLM) and Masked Image Modeling (MIM).
    Additionally, LayoutLMv3 is pre-trained with a Word-Patch Alignment (WPA) objective to learn cross-modal alignment by predicting whether the corresponding image patch of a text word is masked. ``Seg'' denotes segment-level positions. ``[CLS]'', ``[MASK]'', ``[SEP]'' and ``[SPE]'' are special tokens.
    } \label{fig:layarchitecture}
\end{figure*}

The main idea for LayoutLM family is to do multiple pretraining tasks which closely align the image and text level tokens. For the latest LayoutLMv3 shown in figure \ref{fig:layarchitecture}, we take word level feature as token, divide Image as patch and project in latent sapce and append them to text. Then we do 3 level of pretraining
\begin{enumerate}
    \item Mask Language Modelling with text token
    \item Mask Image Modelling with Image token - reconstruct masked image patches, target tokens latent codes from a discrete VAE
    \item Word Patch Alignment - predict from some text token if corresponding image patch is unmasked
\end{enumerate}
 
\subsection{Tasks}

We next discuss the 2 task in screen understanding. To best of our knowledge this is the first work done to propose such tasks in Screen Understanding.

\begin{enumerate}
    \item Visual Question Answering for on screen context
    \item Automated form filling using previous on screen context
\end{enumerate}

\section{Visual Question Answering for on screen context}

The task for visual question answering is to build a system which can answer questions based on information present on the screen. It is an extractive question answering task.

\begin{figure}[h]
\centering
\includegraphics[width=0.5\linewidth]{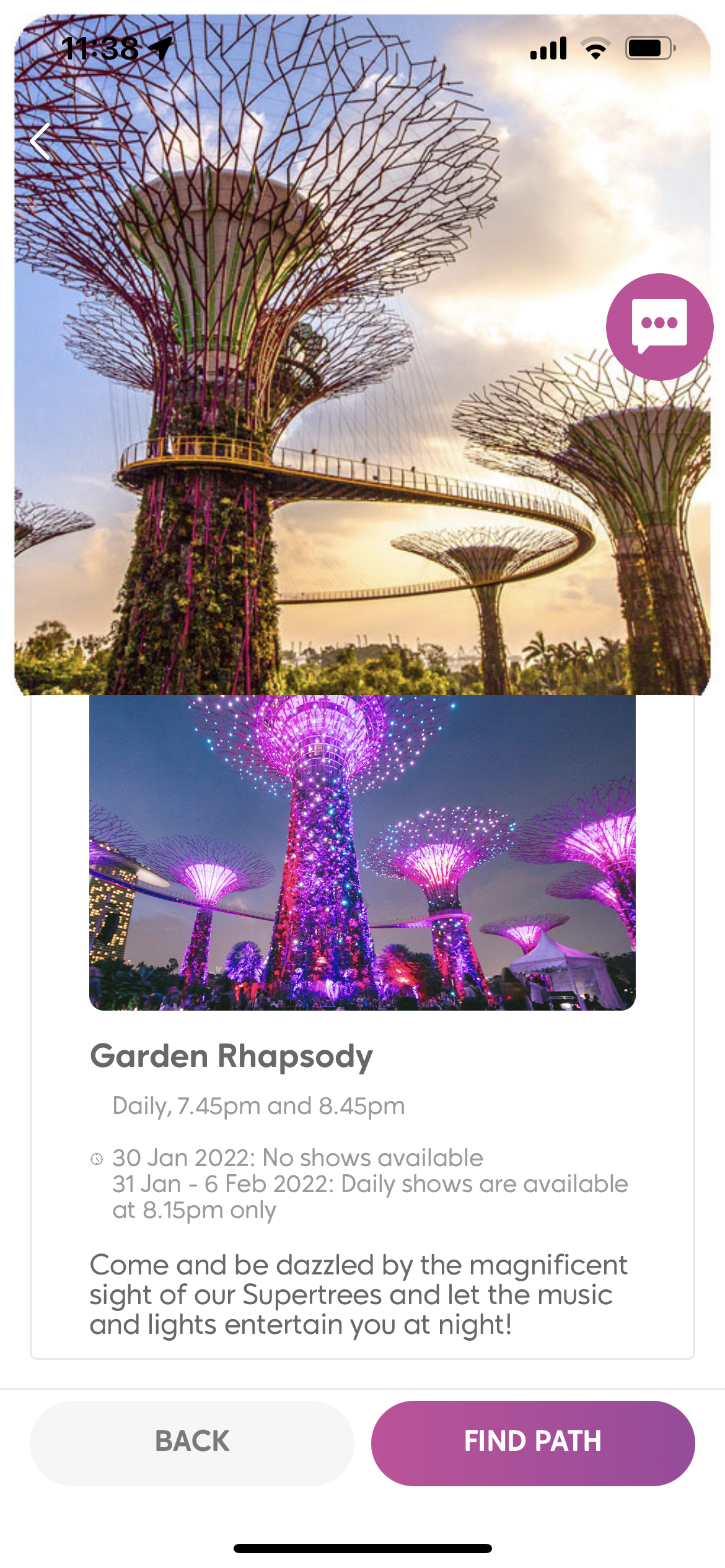}
 \caption{VQA task Problem statement: Build a system which can answer a \textit{natural language query} from a given app view (screenshot + text).\\
E.g. Question (input): When is the daily show?\\
Answer (output): 7:45pm and 8:45pm
}
 \label{fig:layproblem}
\end{figure}

Specifically for a given screen view we need to understand the following to answer a question -
\begin{enumerate}
    \item 'form' -  information is in the form as key:value
    \item 'Layout' - require spatial/layout information like title, heading
    \item 'table/list' - question requires understanding of a table or a list
\end{enumerate}

\subsection{Data and Challenge}

The issue with building a system which works for Visual Question Answering should work for the large variety of apps on appstore. Hence we need need lot of diverse screenshots. We internally collected data from 100,000+ screens from more than 4,500 top-downloaded iOS apps.

\begin{figure}[h]
\centering
\includegraphics[width=0.9\linewidth]{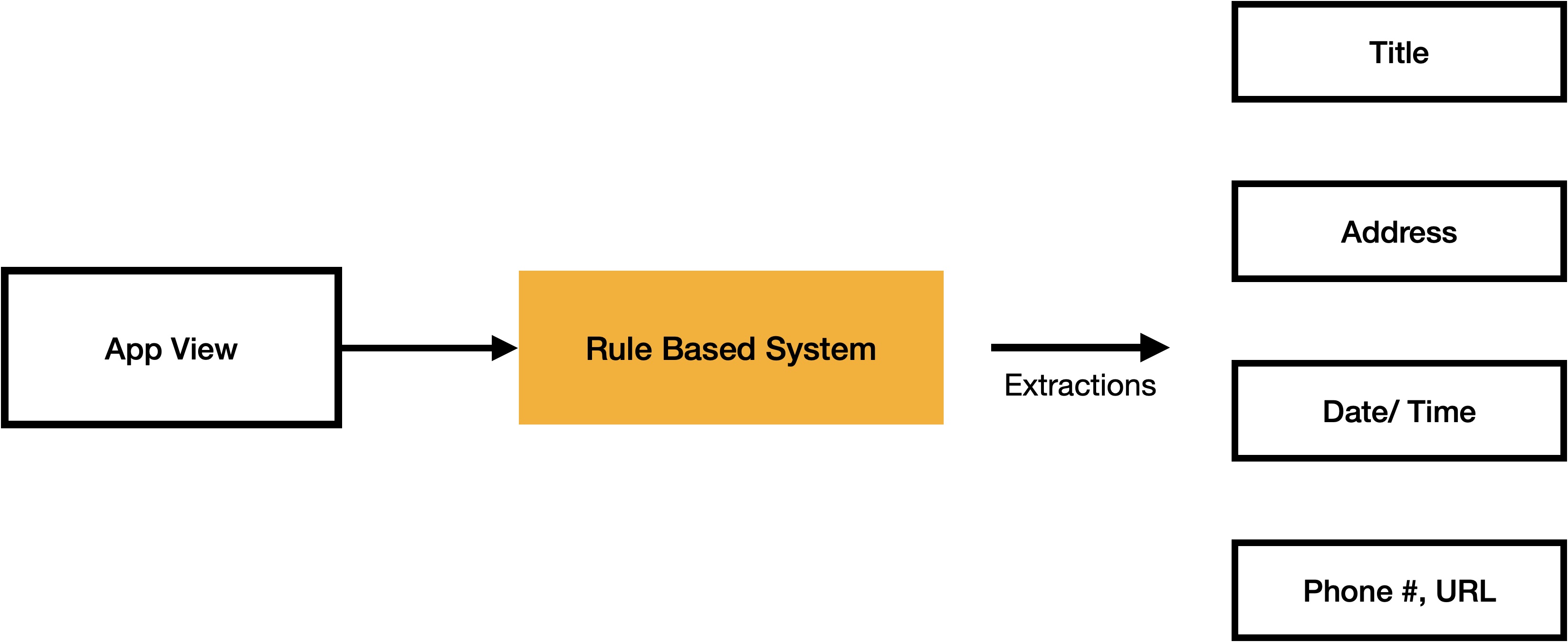}
 \caption{\textbf{Label generation step 1} via extracting predefined data types (aka values)
}
 \label{fig:layrules}
\end{figure}

But the challenge was this was unlabelled i.e. we need correct question, answer pair for a given app view, which we didn't have. To solve this issue we used a 2 step solution. First, we leveraged a rule based system which given an input app view could extract specific data types like date, time, url,address, title (aka values) via rules, as shown in figure \ref{fig:layrules}.

Second, we then found the \textbf{nearest parent text element} that can be approximate as the key for extracted value. Now we can phrase questions in the following format

\textit{Question: what is this {key}?
}

\textit{Answer: {value}}

\begin{figure}[h]
\centering
\includegraphics[width=0.7\linewidth]{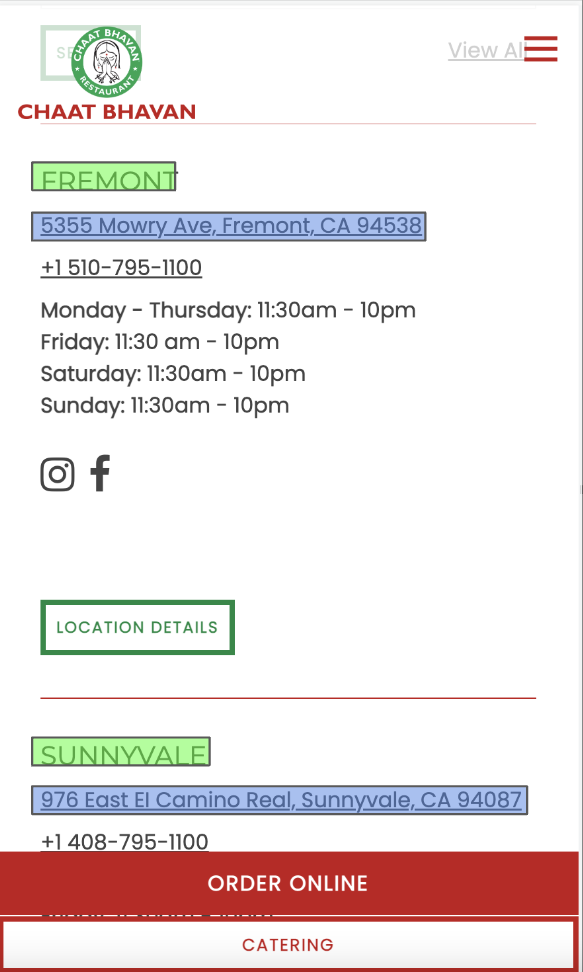}
 \caption{Step 2 of label generation, finding nearest parent text element, and frames questions based on the same. Here we could extract 2 addresses and then had 2 questions in training data.\\
 \textit{Question 1: what is the fremont address?}\\
Answer: 5355 Mowry Ave, Fremont, CA 94538\\
\textit{Question 2: what is the sunnyvale address?}\\
Answer: 976 East El Camino Real, Sunnyvale, CA 94087
}
 \label{fig:laykey}
\end{figure}

See figure \ref{fig:laykey} where we could extract 2 addresses and then had 2 questions in training data.

\subsection{Training Pipeline}

Our training methodology had 4 stages. 
\begin{enumerate}
    \item We start with the LayoutLMv3 model which has been pre-trained on a large amount of open source scanned documents.
    \item Add Question answering (QA) head on top LayoutLMv3. The model is then trained on DocVQA  dataset.
    \item Generate weak labels (question answer pairs) using Rule based system on the top apps dataset. to form a for training \item Finetune on internal apps dataset using incremental learning.
\end{enumerate}

We use DocVQA \citep{mathew2021docvqa} dataset to initialize QA head, because the question answer set of DocVQA is very close to our internal apps dataset training. Further DocVQA labels are clean and gold standard, while our labels are noisy. 

\begin{figure*}[ht]
    \centering
    \includegraphics[width=0.9\linewidth]{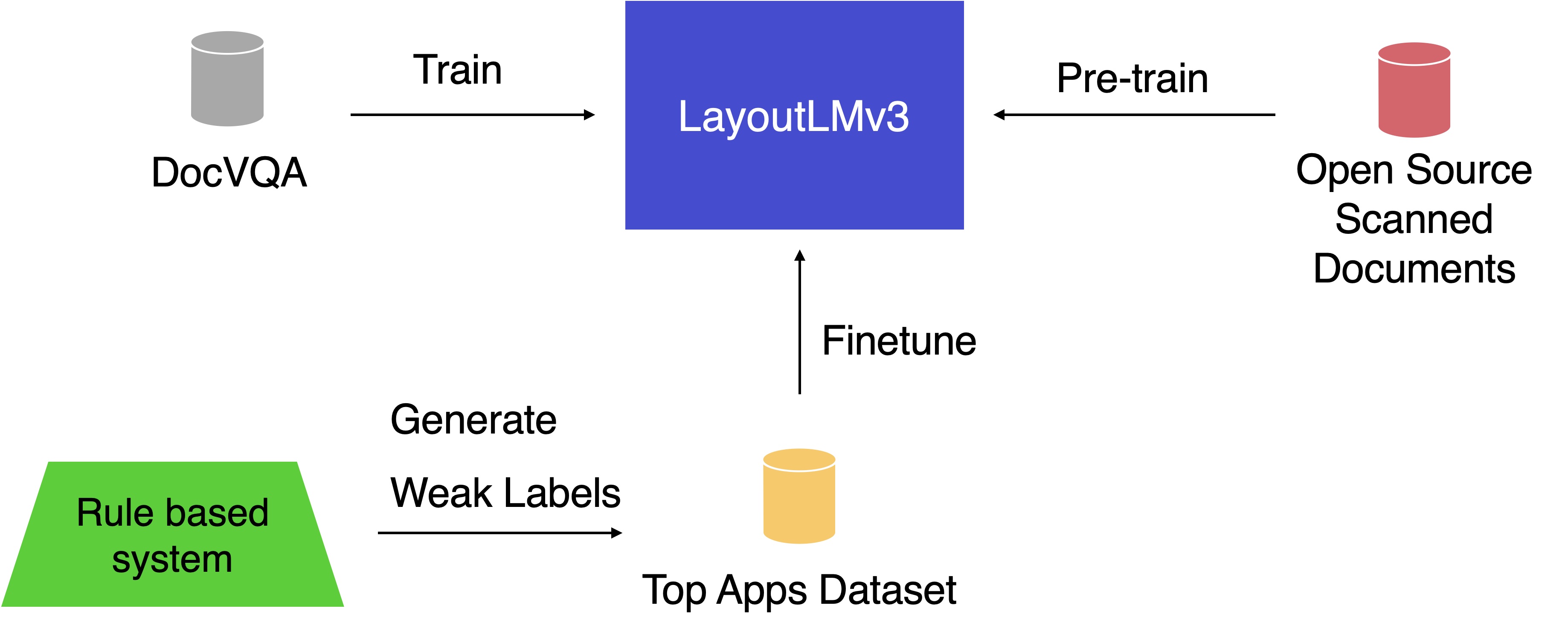}
    \caption{
    \textbf{Training pipeline} (1) Start with the pretraining task of layoutLMv3 (2) Add Question answering (QA) head on top LayoutLMv3 (3) Initialize training of QA head on DocVQA dataset (4) Finetune on weak label generated internal apps dataset using incremental learning.  
    } \label{fig:laypipeline}
\end{figure*}

\clearpage
\subsection{Question Types}
The following 6 types of questions were generated using the weak labelling procedure described earlier on tops apps dataset used for final training. Note if we found that a value has an associated key, we used the key as in forming question by replacing \{\} with the key. We call these questions keyed questions as highlighted in blue. Two examples of keyed questions are shown in figure \ref{fig:laykey}

\paragraph{title}
"What is the document about?", "What is the title?", "What is it about?"

\paragraph{phone number}
"What is the phone number?", "What is the number?"\\
\textit{\color{blue}{"What is the \{\} phone number?", "What is the \{\} number?"}}

\paragraph{email}
"What is the email?", "What is the email address?"\\
\textit{\color{blue}{"What is the \{\} email?", "What is the \{\} email address?"}}

\paragraph{url}
"What is the url?", "What is the link?"\\
\textit{\color{blue}{"What is the \{\} url?", "What is the \{\} link?"}}

\paragraph{address}
"What is the address?", "What is the location?"\\
\textit{\color{blue}{"What is the \{\} address?", "What is the \{\} location?"}}

\paragraph{'DateTime'}
"What is the date?", "What is the time?", "When is it?"\\
\textit{\color{blue}{"When is the \{\}?", "What time is \{\} scheduled?", "When is \{\} scheduled?", "What date is \{\} scheduled?"}}

\subsection{Observations}

We now analyze the fine-tuning on our top apps dataset starting from DocVQA checkpoint. We observe the following trends.

\paragraph{Observation 1}
The validation loss hasn’t dropped much even though train loss has decreased. Similarly validation f1 hasn't increased much over training.

\begin{figure}[h]
\centering
\includegraphics[width=\linewidth]{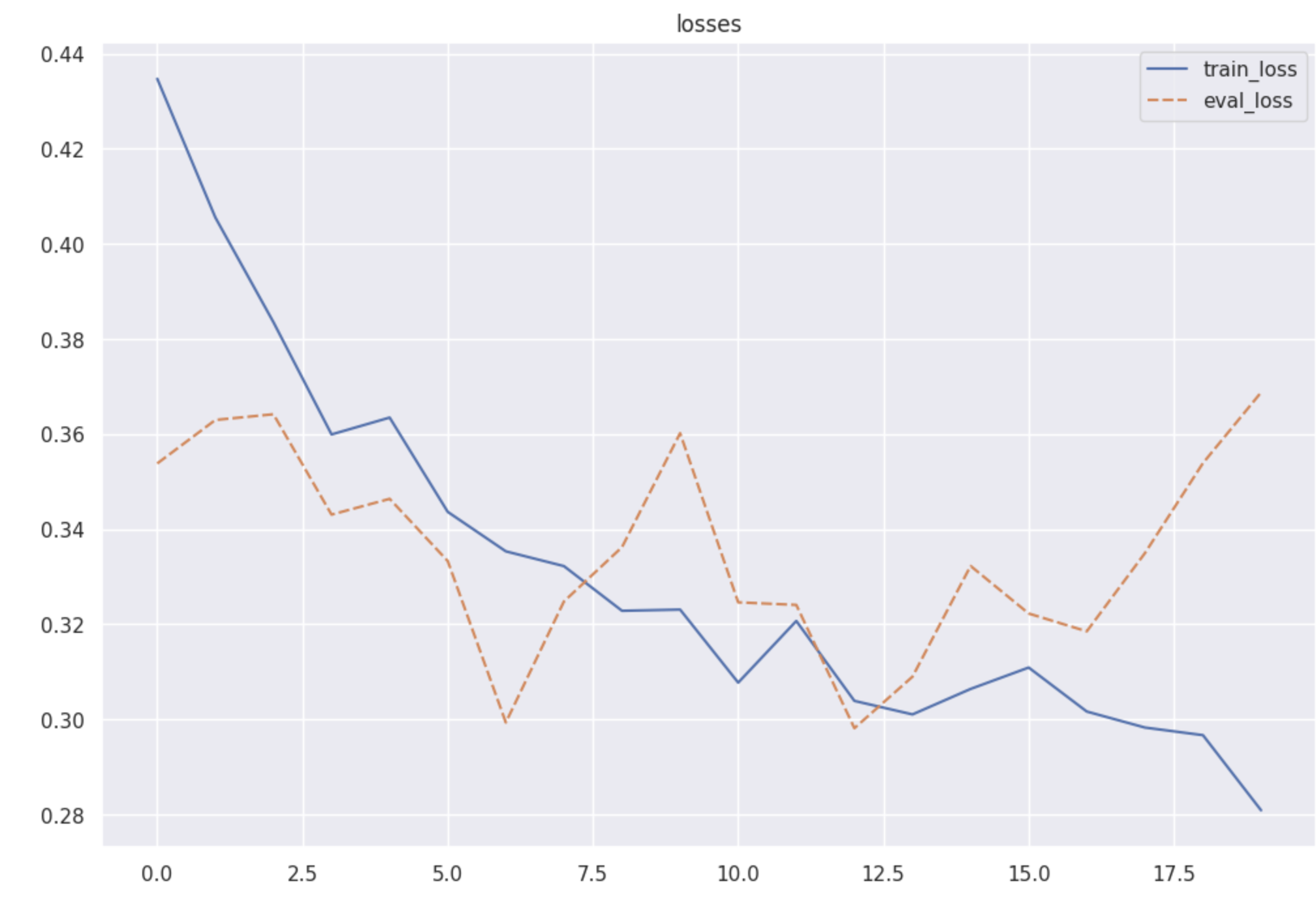}
 \caption{Overall loss curve while fine-tuning on the tops apps dataset starting from the DocVQA train checkpoint.
}
 \label{fig:layoverallloss}
\end{figure}

\begin{figure}[h]
\centering
\includegraphics[width=\linewidth]{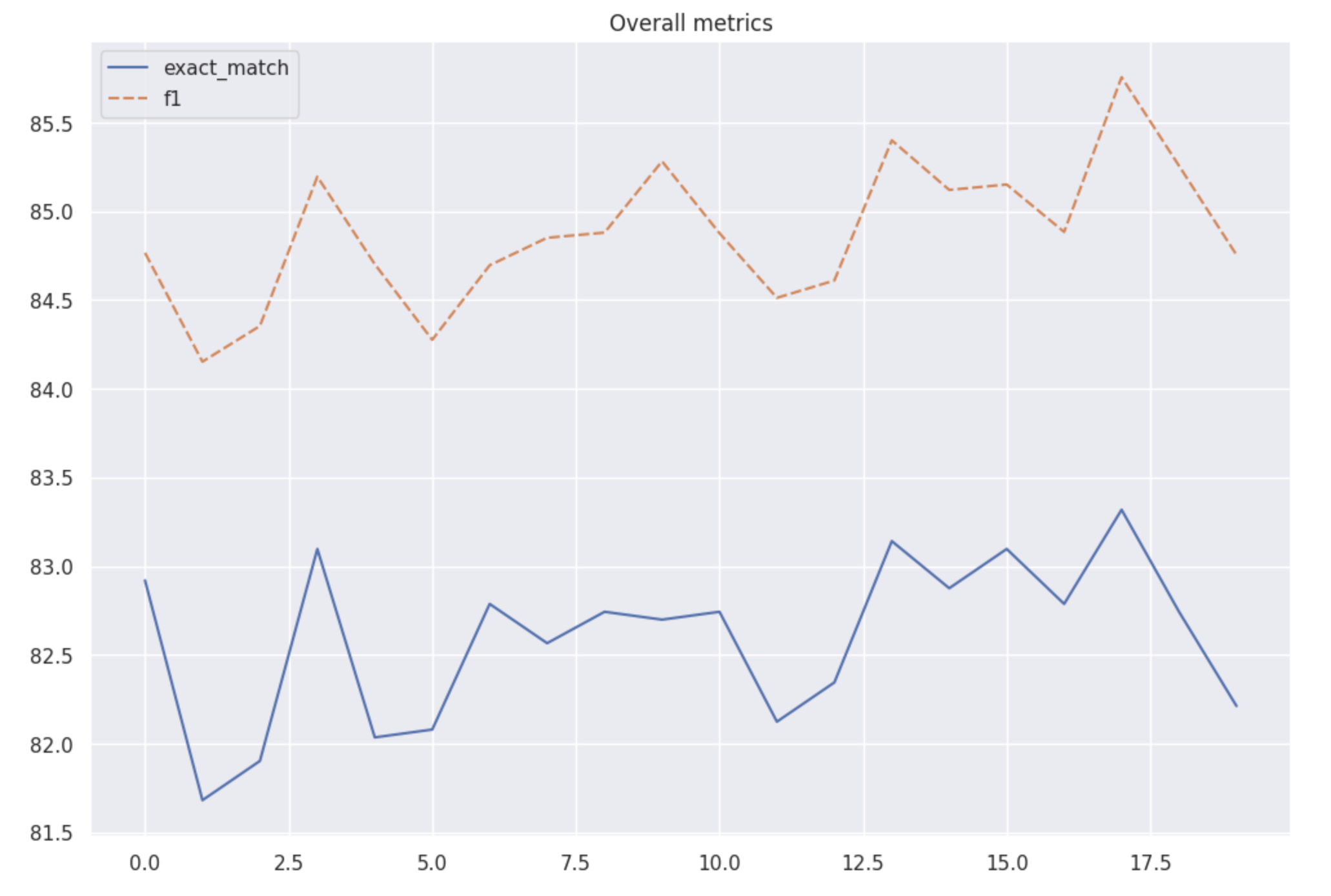}
 \caption{Overall metrics f1 and recall while fine-tuning on the tops apps dataset starting from the DocVQA train checkpoint.
}
 \label{fig:layoverallmetrics}
\end{figure}

\paragraph{Possible Reasons}

\begin{figure}[h]
\centering
\includegraphics[width=0.7\linewidth]{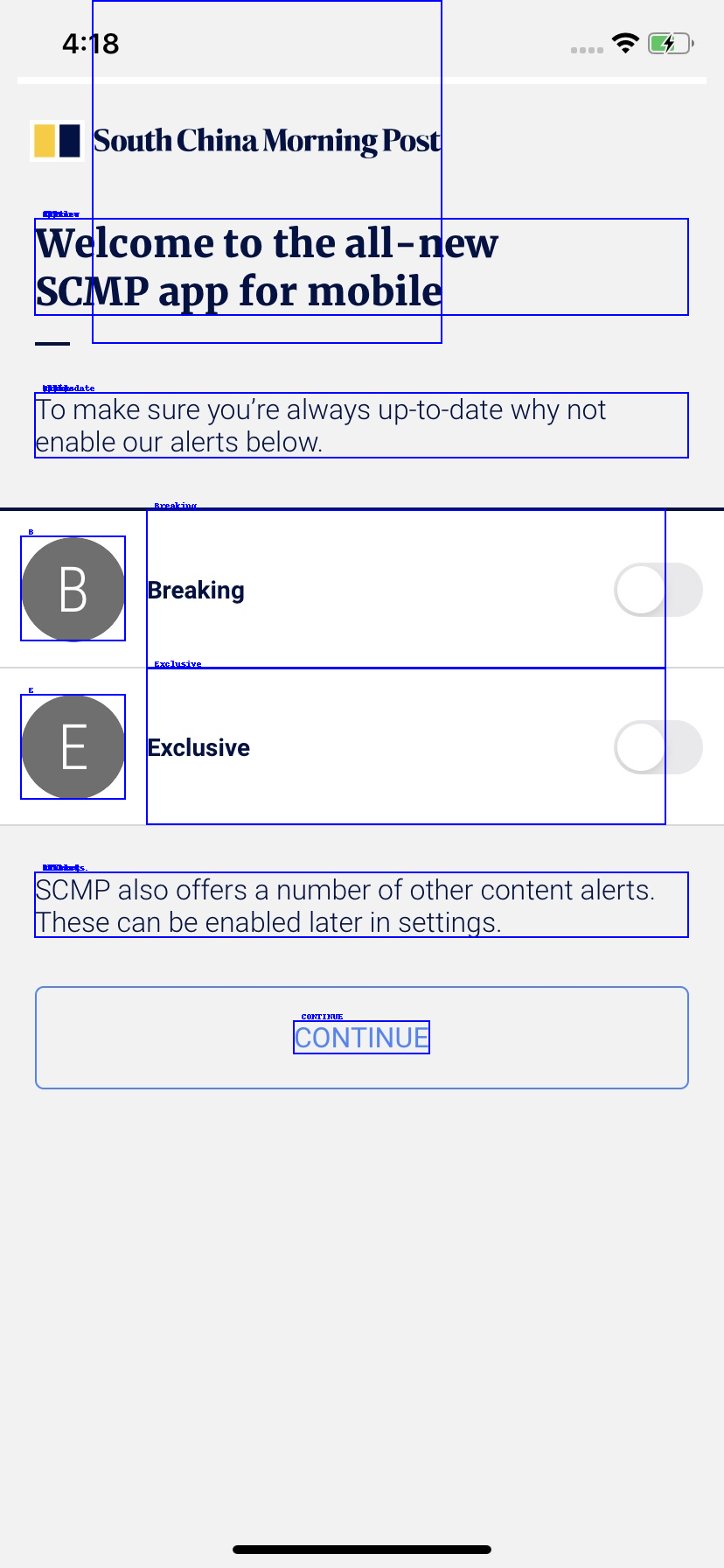}
 \caption{An example of issue present in tops apps dataset where the input bounding boxes are not so tight.}
 \label{fig:laybboxissue}
\end{figure}

\begin{enumerate}
    \item We start fine tuning the model from the trained checkpoint of DocVQA database (50,000 questions defined on 12,000+ document images) which is already a good starting point.
    \item The dataset we used for fine tuning has fews issue - the bounding boxes are not too tight and spill to other text regions as shown in figure \ref{fig:laybboxissue}. Also there are ghost bounding boxes for text in the fine tuning dataset which are not visible to users on screen, thereby confusing the model.
\end{enumerate}

\paragraph{Observation 2}
The validation f1 for title has improved marginally from 98.8 to 99.4.

\paragraph{Possible Reason} The DocVQA dataset has around 500 samples for title, while majority of our train samples were title. This can be estimated from validation data which has 1379 samples for title, validation and train data follow same distribution.

\begin{figure}[h]
\centering
\includegraphics[width=\linewidth]{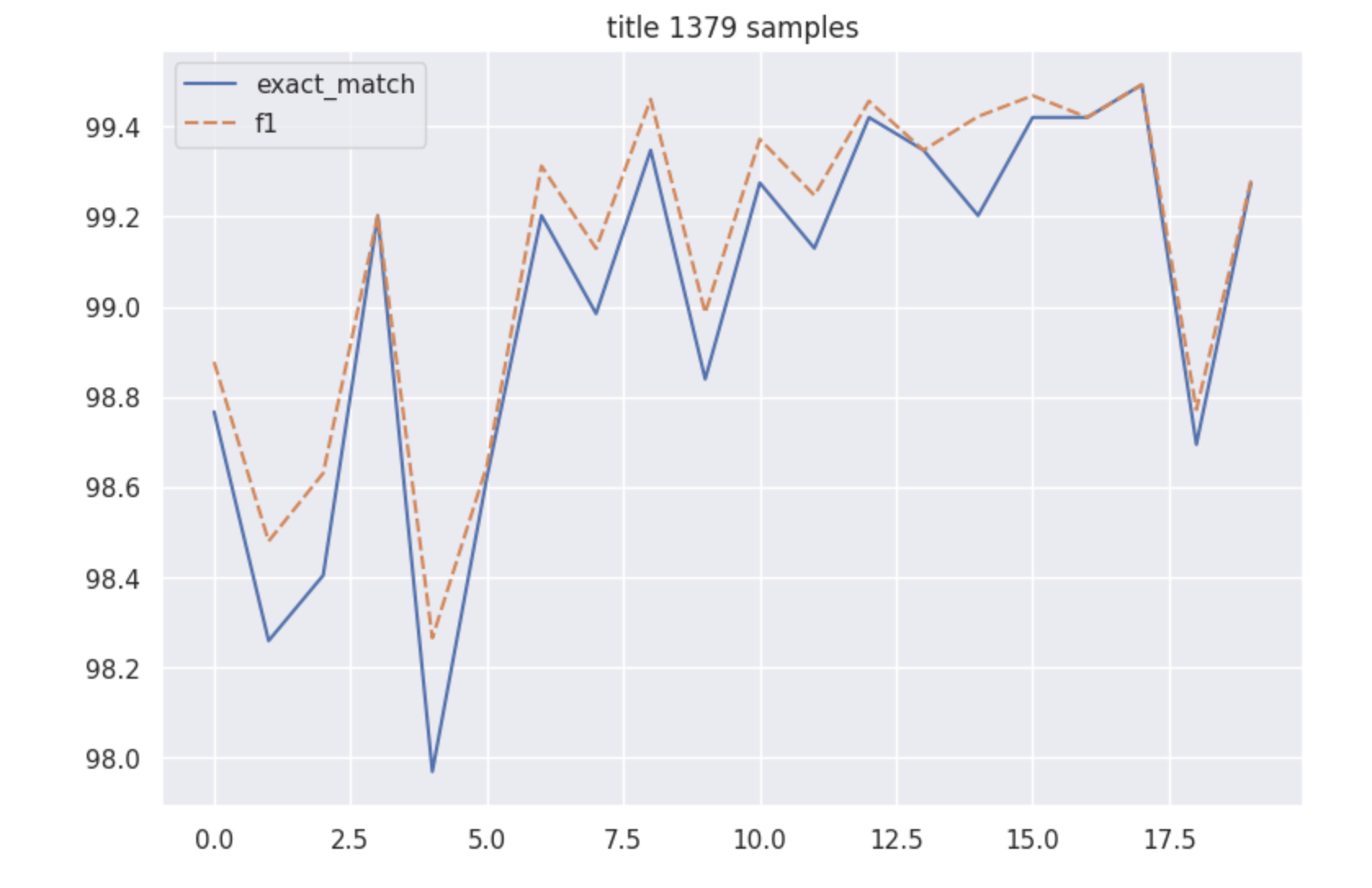}
 \caption{The validation f1 for title has improved marginally from 98.8 to 99.4.
}
 \label{fig:laytitle}
\end{figure}

\paragraph{Observation 3}
The validation f1 for date has even fact dropped after our training from 50 to 42.

\paragraph{Possible Reason} The DocVQA dataset has over 4,500 samples for date, while we had only very limited datetime samples in training. This can be estimated from validation data which has 61 samples for datetime, validation and train data follow same distribution.

\begin{figure}[h]
\centering
\includegraphics[width=\linewidth]{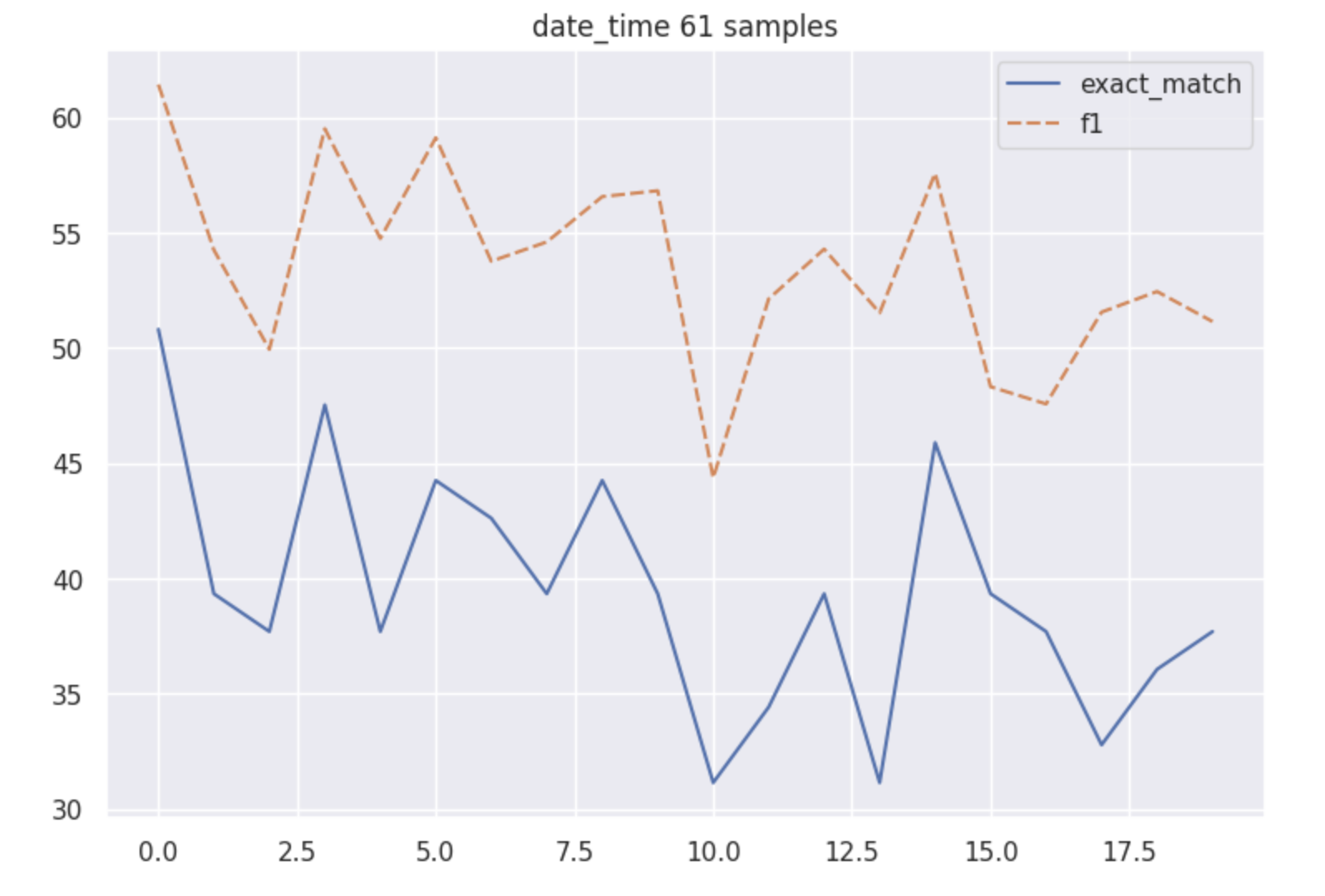}
 \caption{The validation f1 for date has even fact dropped after our training from 50 to 42.
}
 \label{fig:laydate}
\end{figure}

\subsection{Results}

\begin{figure*}
    \centering
    \includegraphics[width=0.9\linewidth]{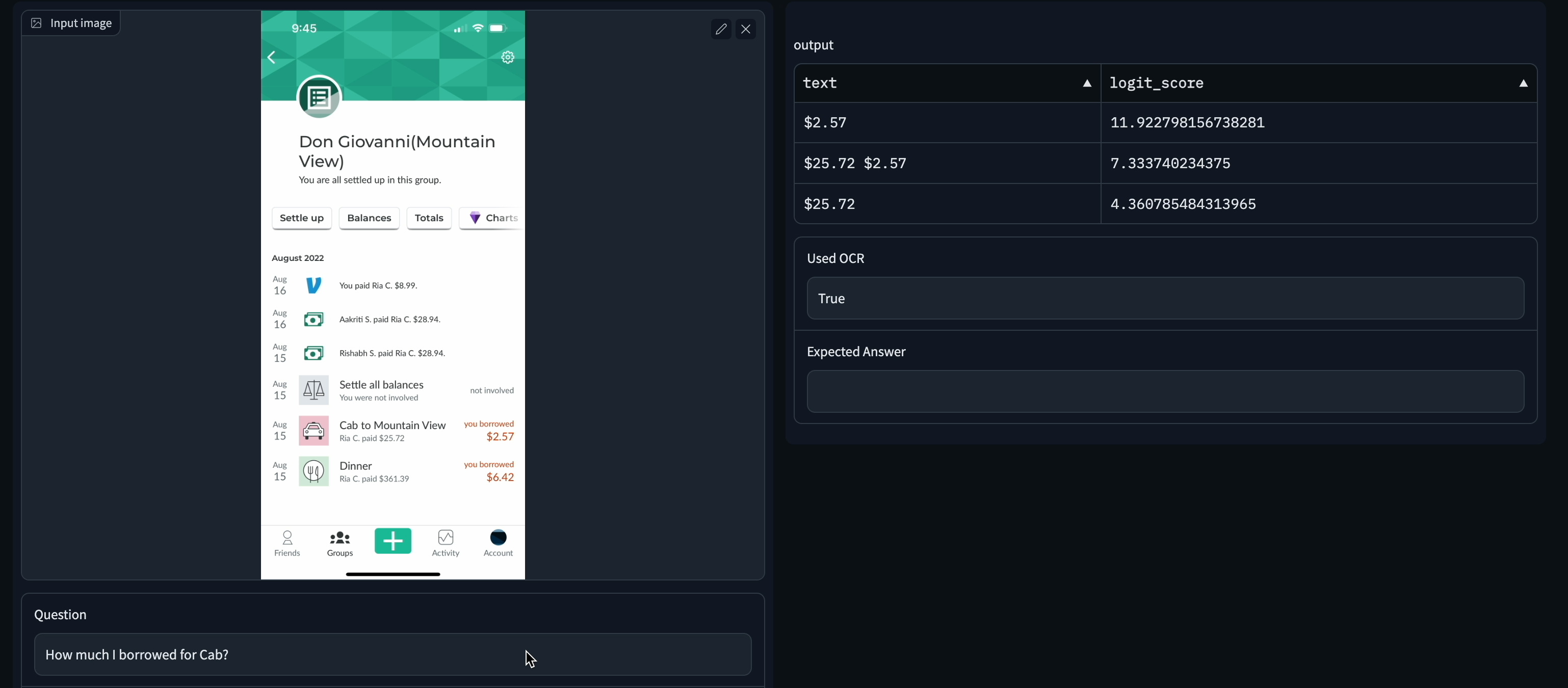}
    \caption{The model is able to effectively do layout based understanding and do "I" $\rightarrow$ "you" association to answer the question about price of cab. We ask the question “How much I borrowed for Cab?”, model predicts the rightly predicts correct answer of \$2.57
    } \label{fig:layrescab}
\end{figure*}

\begin{figure*}
    \centering
    \includegraphics[width=0.9\linewidth]{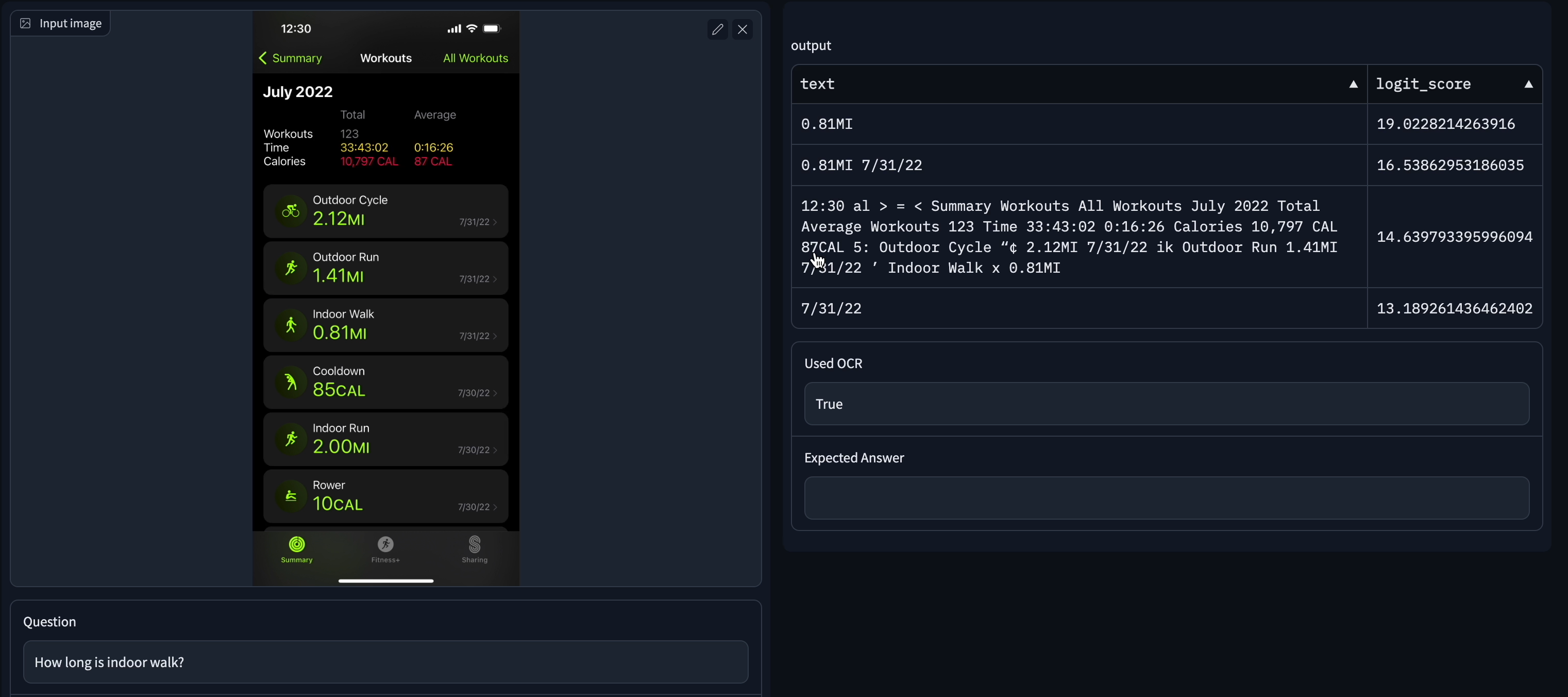}
    \caption{The model is able to effectively predicts unseen generic data types like length without explicitly being trained on it. We ask the question “How long is indoor walk?” the model predicts the correct answer of 0.81M which is a data type model hasn’t seen before
    } \label{fig:layresrun}
\end{figure*}

\begin{figure*}
    \centering
    \includegraphics[width=0.9\linewidth]{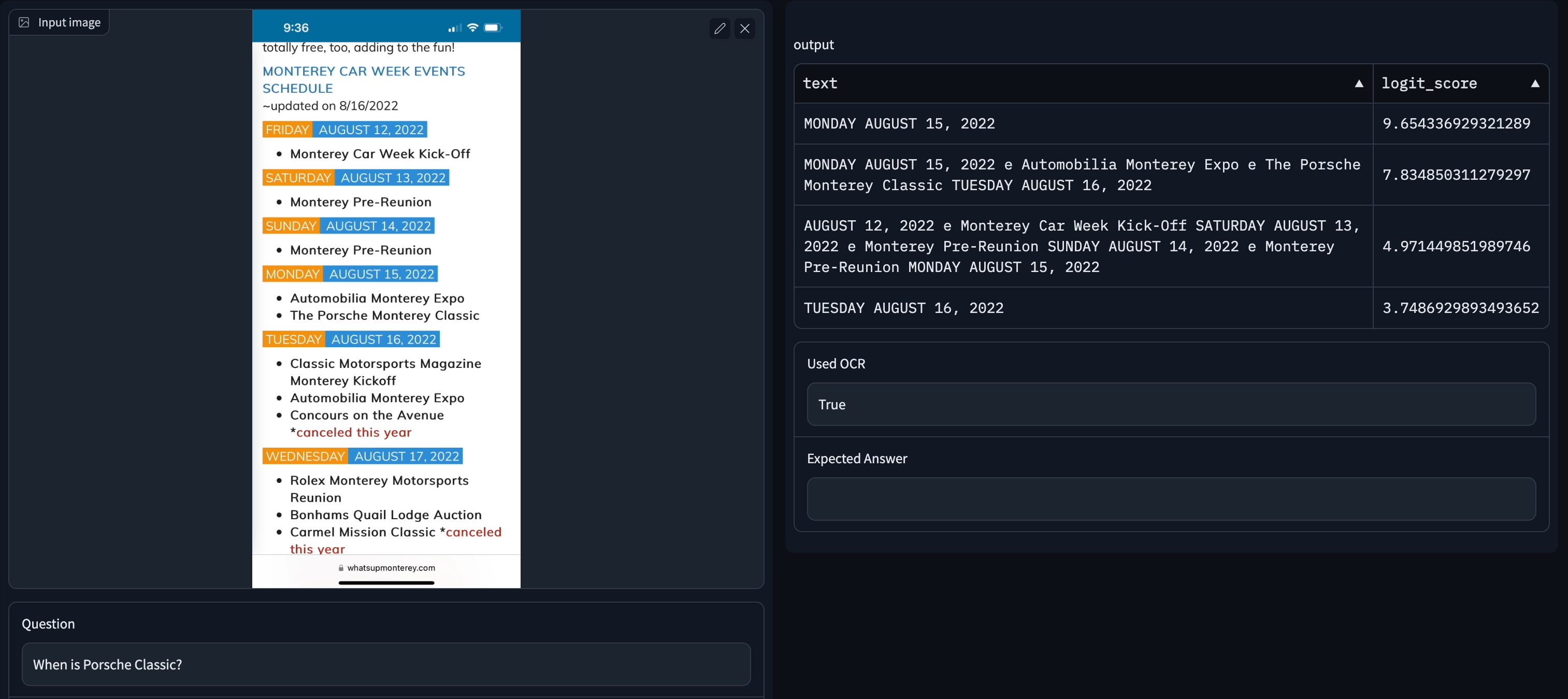}
    \caption{The model is able to rightly parse a list in the usual semantic way and do robust entity association to associate Porsche classic with correct date of August 15 even thought spatially August 16 lies more close to the text.
    } \label{fig:layreslist}
\end{figure*}

\begin{figure*}
    \centering
    \includegraphics[width=0.9\linewidth]{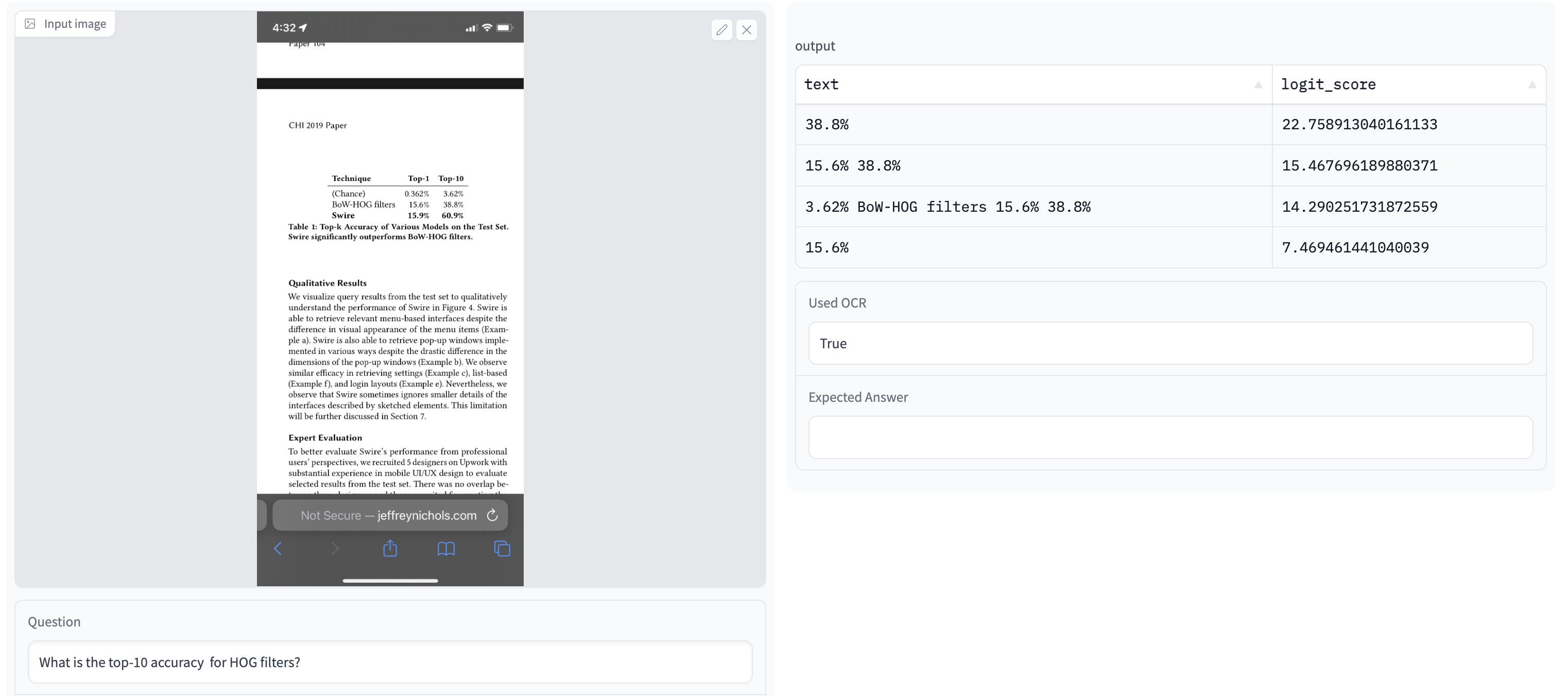}
    \caption{The model is effectively able to answer queries for tables, understand the structure of tables to map rows and columns. We ask the query “What is the top-10 accuracy for chance?” the model predicts right answer of 3.62\%.
    } \label{fig:layrestable}
\end{figure*}

Below we show 4 examples in figures \ref{fig:layrescab}, \ref{fig:layresrun}, \ref{fig:layreslist}, \ref{fig:layrestable}

Each example highlights a particular challenge the model is able to solve without being explicitly trained on those samples.

Examples include
\begin{enumerate}
    \item Understand natural query and do structure based association in the cab example.
    \item Work on generic new data types like length.
    \item Do robust entity association in a list shown in the car race schedule example.
    \item Understand tables
\end{enumerate}

\subsection{Limitations}

The following limitations were noticed while working on the model

\begin{enumerate}
    \item The proposed model can't understand and reason about images. The model LayoutLM can understand the structure and layout but it can’t reason nor understand if there is an image of a dog. This is because understanding the image was not part of its training objective, nor does it have any image-based backbone?
    \item The model doesn't do intent classification of query (e.g. model that the question like "How long was outdoor run?" expects length as an answer) so the generated answer can be arbitrarily bad, and may not be even what user expects (e.g. model may predict date when asked about the same length question if a date is nearby the text "outdoor run" in the input screen.)
    \item The \textbf{latency} of the model is quite large around 450 ms with 343M parameter. To make it work on device in real time, we have to shrink its runtime to less than 50M parameters.

\end{enumerate}

\subsection{Future Directions}
\begin{figure}[h]
\centering
\includegraphics[width=0.7\linewidth]{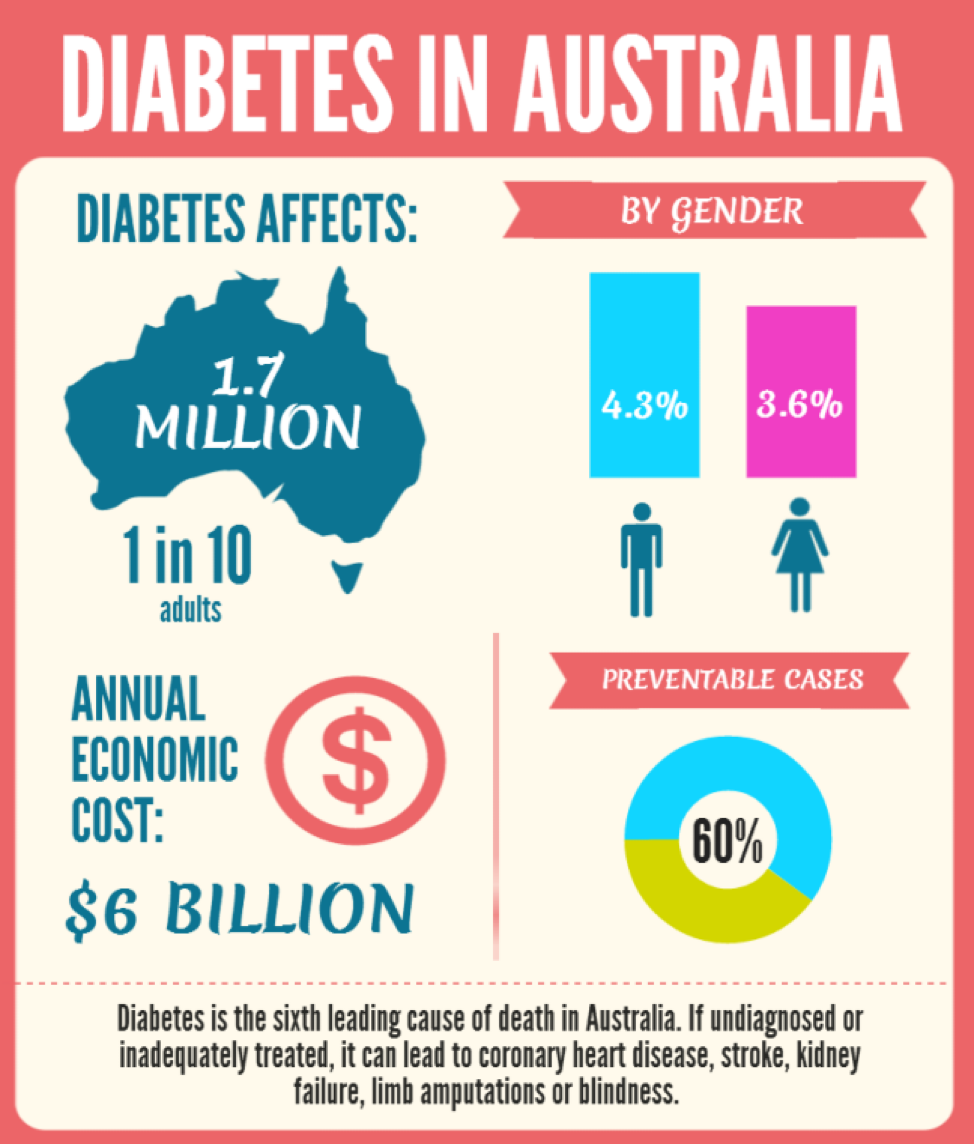}
 \caption{Support for infographics understanding for answering questions\\
 \textit{How many females are affected by diabetes?} 3.6\%\\
\textit{What percentage of cases can be prevented?} 60\%\\
\textit{What could lead to blindness or stroke?} diabetes
}
 \label{fig:layinfo}
\end{figure}

\begin{enumerate}
    \item One possible future direction for VQA is to look for predicting the bounding box of evidence when predicting an answer.
    \item Support for Infographics/ Charts understanding. E.g. given an image Answer question which require understand image, and diagrams. As shown in figure \ref{fig:layinfo}

\end{enumerate}

\begin{figure}[h]
\centering
\includegraphics[width=0.9\linewidth]{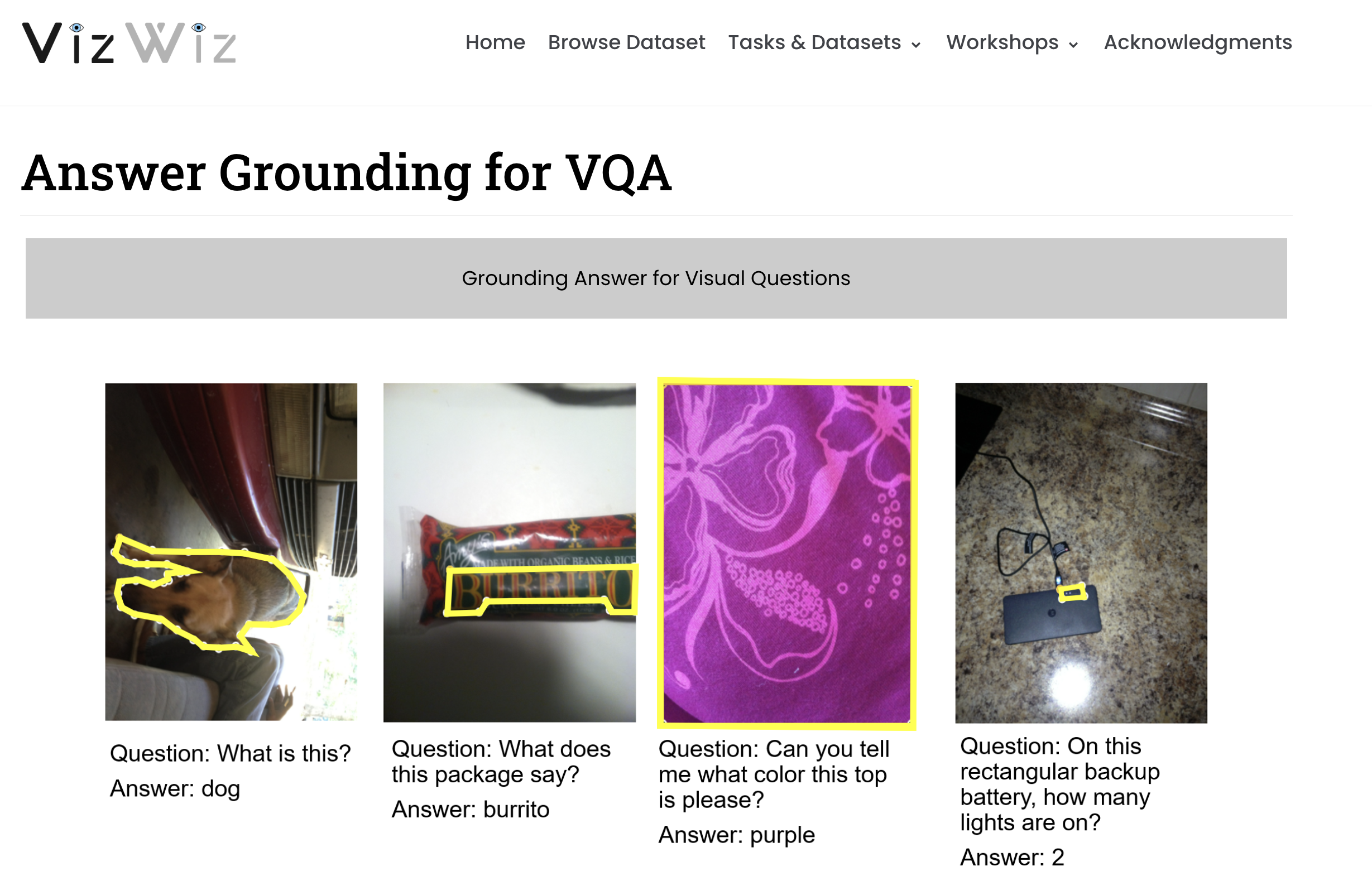}
 \caption{vizwiz dataset with visual grounding for Question answering.
}
 \label{fig:layviz}
\end{figure}

\clearpage

\section{Automated form filling using previous on screen context}

The task of automated form filling refers to the task when the user is filling a form and information required to fill the same exists in one of the recent screens user previously visited. Currently the user requires to go back and forth between screens, copy each information individually and then paste that information in the relevant fields one by one. This is a slow and repetitive process on user end. A way which automates the form filling process for user by suggesting the information from previous screen automatically relevant to the user in the current form.

The form filling suggestions can also made when the user is looking to input an information present in previous screen the user visited. To the best of our knowledge, this task has not been formally defined nor research previously.

\subsection{Data and challenge}

Since the task didn't exist before we need to create our own dataset for the task. We collect around 150 samples of different forms including flight reservations, hotel reservations, ticket creation and screenshot of previous screens. We also collect information of screen where the user is in a chat with an agent and requested some information present on previous screen in the current chat.

\begin{figure}
\centering
\includegraphics[width=1.0\linewidth]{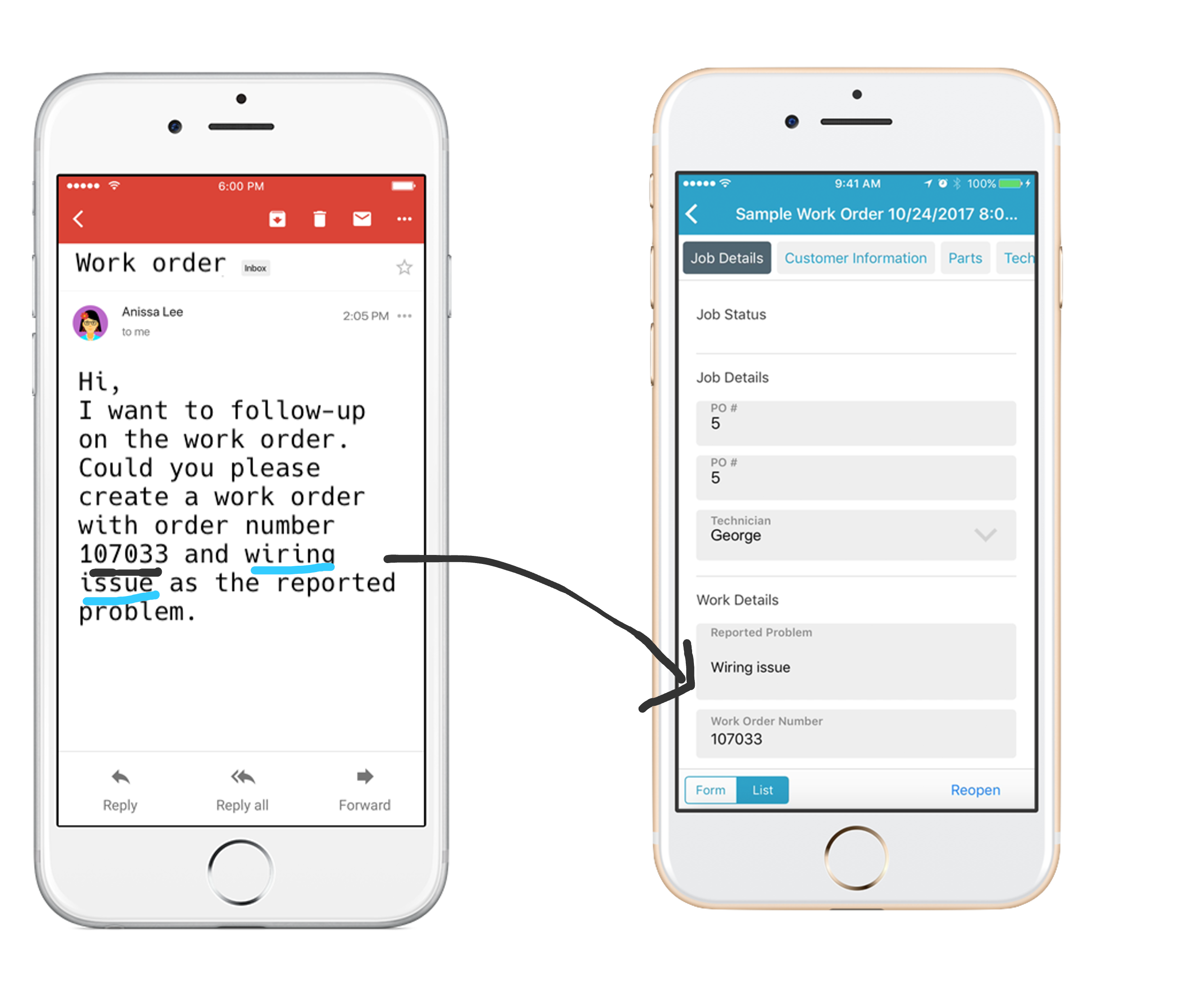}
 \caption{An example of automated form filling where the input form on the right is auto-filled based on information in the left screen.}
 \label{fig:formfill}
\end{figure}

An example is shown in figure \ref{fig:formfill} where we have 2 questions about "Reported problem" and work order number which can filled from information in the previous screen.

The challenge here is to design a system which process each screen of user once, and stores an intermediate representation. We can then use multiple such screen representations to find appropriate form fields which could be filled, without having to recompute the intermediate representation every time.

\begin{table}
\begin{tabularx}{\linewidth}{|l|X|}
\hline
Form Screen     & Flight, hotel reservations, ticket creation                               \\\hline
Information Screen        & Email, Chat , webpages                          \\\hline
Total pairs & 152\\\hline
Avg Questions-answer & 2.3 per pair\\ \hline
\end{tabularx}
\caption{Information about training data for form filling}
 \label{tab:formdata}
\end{table}

\subsection{Pipeline}

The pipeline consists of keeping a buffer of previously visited screen and the current screen. First we pass the current screen and previous screen through LayoutLMv3 individually. Then the last layer representations are concatenated passed to a question extraction head which extractions spans of relevant questions which can be answered in the current screen (form screen) from the previous screen (info screen). The extracted question tokens along with representation of info screen are then passed to an answer extraction head which predicts the answer span for each of the extracted tokens. 

Note we use the same LayoutLMv3 model instance to extract representations for both form screen and info screen, which the representations are then differentiated when based to respective question or answer extraction head.

The latency of the model is increased by the faced that for each extracted question we need to run the answer extraction head to get span of answer predicted for a particular question.

Note that both question answering and answer extraction head are 3 layer feedforward network with a cross attention layer at the start.

\subsection{Results}

Since we have 2 additional heads to train on top of LayoutLM we start evaluate each of the units individually as shown in table \ref{tab:formres}. We observe few trends -

\paragraph{Observation 1}
The question extraction is an easier task than answer extraction for a given question. 

\paragraph{Possible Reason} This may be related to extra cues from image space associated with empty blank space which helps in easy classification of question.

\paragraph{Observation 2}
Recall is usually higher than precision for both the question and answer extraction task.

\paragraph{Possible Reason} This may be attributed to fact that our model is able to retrieve back most questions and answers albeit with extra tokens around it leading to lower precision than recall.

\subsection{Limitations}

The following limitations were noticed while working on the model

\begin{enumerate}
    \item The latency of LayoutLMv3 is still very high to deploy on mobile in real time.
    \item We need to run answer extraction head for each question we find. This increases our run time.
    \item The system is currently only trained for a small sample of data and larger level study needs to be done to see its effectiveness.
\end{enumerate}

\begin{table}[]
\begin{tabularx}{\linewidth}{|l|l|X|}
\hline
Task   & F1 score &  Recall            \\\hline
Question extraction        & 74.50 & 78.32                          \\\hline
Answer extraction & 63.87 & 68.12\\\hline
\end{tabularx}
\caption{Testing data results on form fillings}
 \label{tab:formres}
\end{table}

\begin{figure*}
    \centering
    \includegraphics[width=0.9\linewidth]{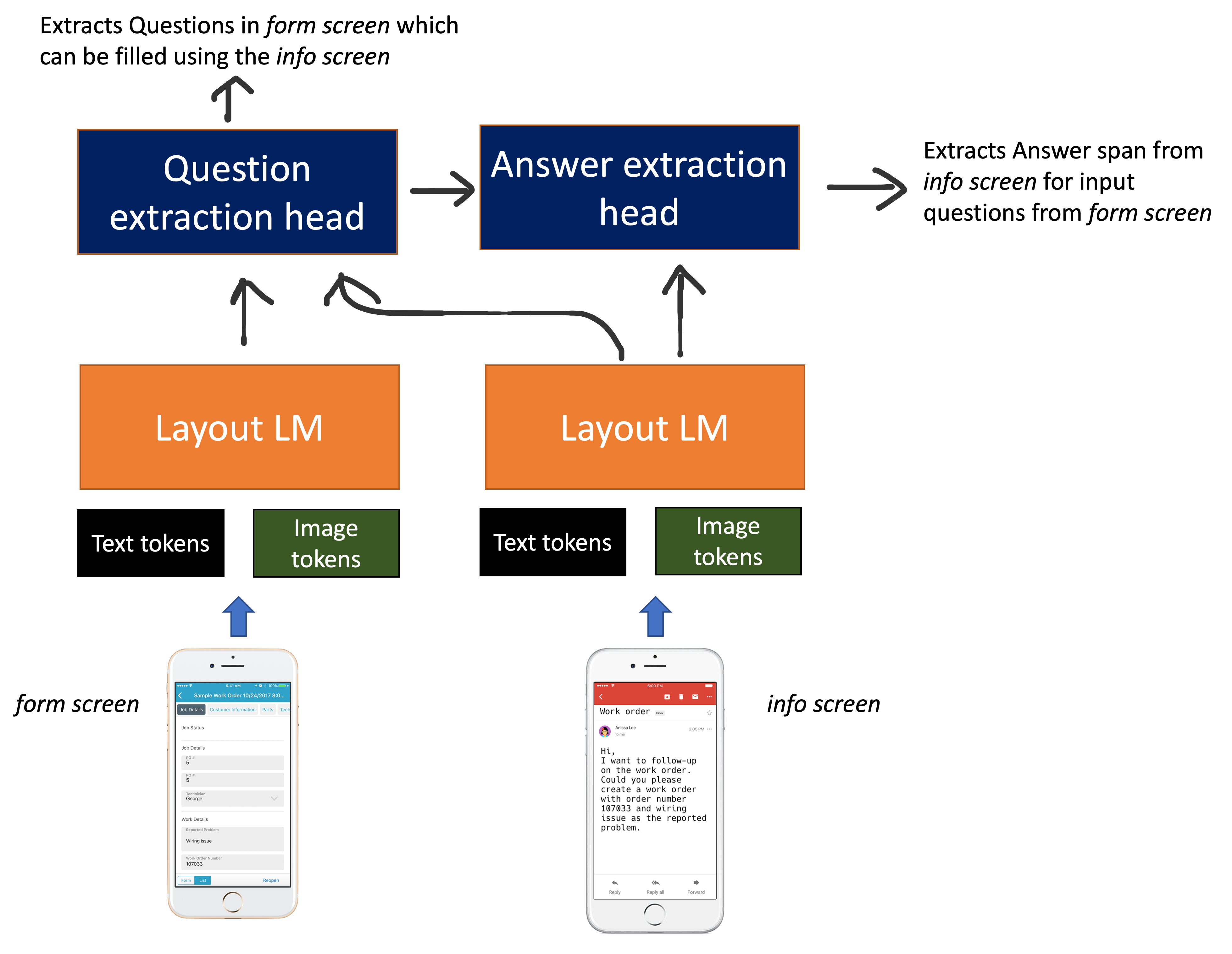}
    \caption{
    \textbf{Pipeline} (1) We pass each the current form screen and previous info screen to layoutLMv3 individually. (2) the last layer representations are concatenated and passed to a question extraction head which extractions spans of relevant questions which can be answered in the current screen (form screen) from the previous screen (info screen). 
    (3) The extracted question tokens along with representation of info screen are then passed to an answer extraction head which predicts the answer span for each of the extracted tokens. 
    } \label{fig:laypipeline2}
\end{figure*}

\clearpage

\section{Smart Replies}

\subsection{Introduction}

Smart replies are automated generated short responses to email or chat in a conversation especially for a phone application, which assists a user to quickly respond to large variety of messages which require similar response. This is meant to save the characters a user is supposed to type on a mobile device and hence save time.

\begin{figure}[h]
\centering
\includegraphics[width=0.9\linewidth]{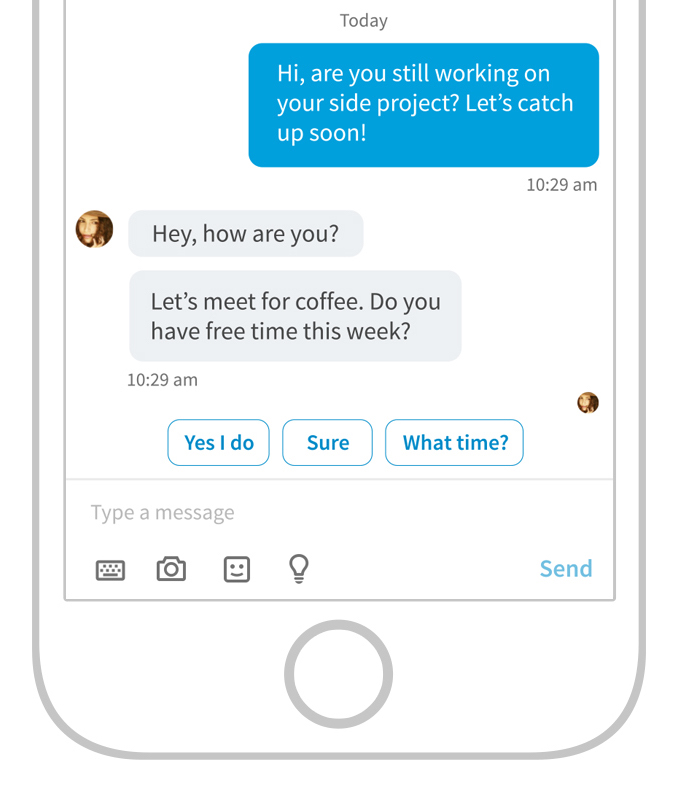}
 \caption{Smart replies model suggests relevant responses in a conversation history to the user.
}
 \label{fig:smartcomv}
\end{figure}

\subsection{Related Work}

The earliest smart reply was for Gmail \citep{kannan2016smart} and used a Seq-to-Seq model to encode a message and then decode a response. To ensure only relevant emails get generated responses and the authors proposed a classifier which based on the email content and its metadata (origin, subject) will screen out marketing emails or emailing requiring more thoughtful longer responses. At the end only about 10-15 \% of the all emails were filtered to be used for further generating smart reply.

To ensure only higher quality messages get suggested as a response, a \textbf{response set} is pre-computed which is the set of all valid responses. And for a given input message, a reply is then searched during decoding only in the valid response set space. To ensure diversification of responses, each response is pre-assigned an intent. Now where searching for a reply to a message, it is ensured that messages which atleast 2 different intents are recommended to user.

\begin{figure}[h]
\centering
\includegraphics[width=0.9\linewidth]{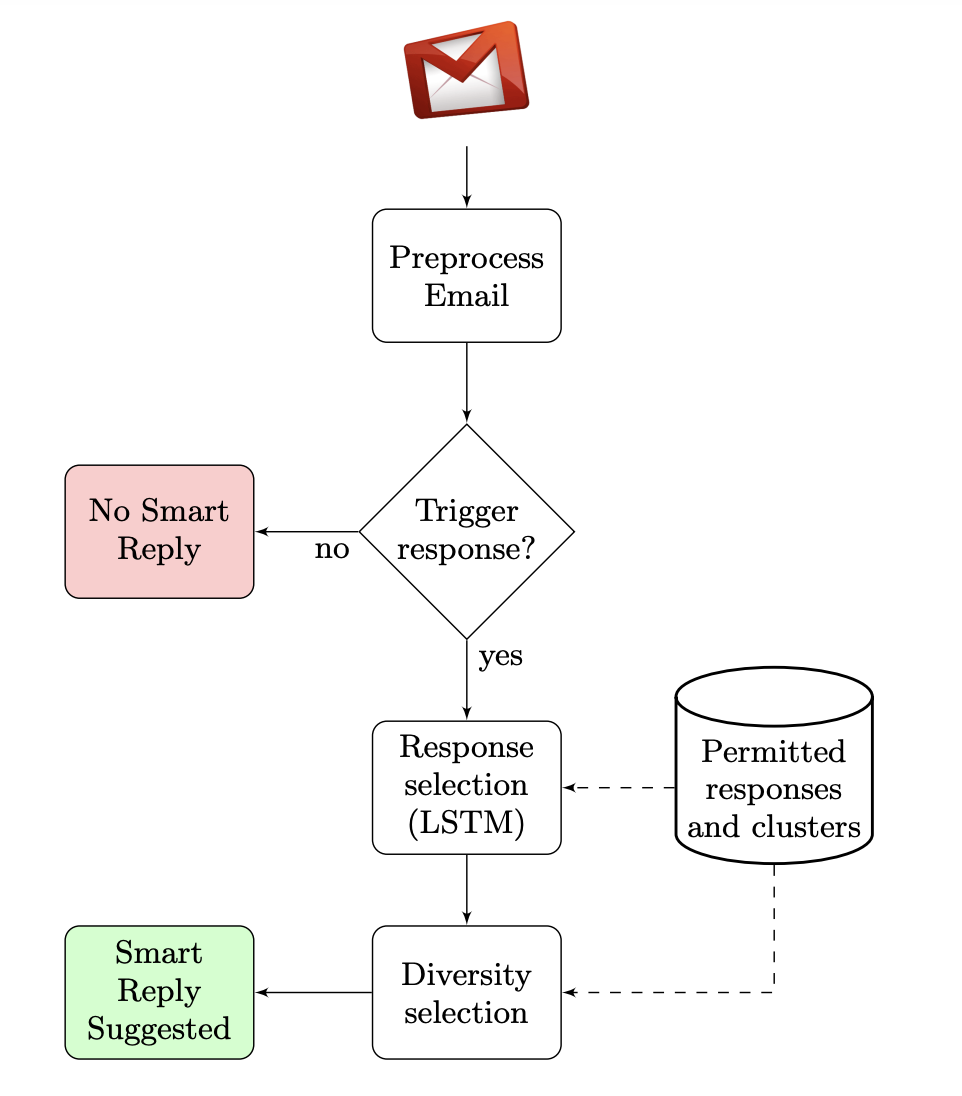}
 \caption{
}
 \label{fig:smartgmail}
\end{figure}

The current approaches \citep{henderson2017efficient} now works on a \textbf{bi-encoder} based approach where first a large set of message-reply (m-r) pairs are collected from the users at a commercial level. The model is then trained on one-on-one message-reply (m-r) pairs from commercial email data. The symmetric loss function is then minimized. It is a modified softmax on dot products between m-r encoding in equation \ref{symloss} where  $s_{i,j}=e^{\phi(m_{i}) \cdot \phi(r_{j})}$.

{\small
	\begin{equation}\label{symloss}
	    p(m_{i}, r_{i})\\ =\frac{s_{i,i}} 
         {\sum_{j}{s_{i,j}} 
         + \sum_{k}{s_{k,i}} 
         - s_{i,i}}
	\end{equation}
During prediction, the authors then compute the matching score ($\cdot$) between the message and pre-computed response set vectors. Then a language-model (LM) penalty is added representing the popularity of responses to bias the predictions towards more common ones. Using this score in equation \ref{inference} the authors first select top $N_1$ responses, and down-select to top $N_{2}$ after deduplication using lexical clustering, before presenting to users.

{\small
    \begin{equation}\label{inference}
        Score = \phi(m_{i}) \cdot \phi_{K}(r_{k})) + \alpha LM_{K}(r_{k})
    \end{equation}
}

\subsection{Tasks}

Smart replies models are currently great for English and high resource language conversations. Further they are only trained for short reply pairs and often the reply lacks diversity. Lastly the smart replies currently don't leverage external knowledge available about certain users during a reply.

To solve these issues we propose 2 new tasks in smart reply space. To best of our knowledge, this is the first work to propose such task.

\begin{enumerate}
\item Suggesting smart replies for multilingual speakers with code switching
\item Suggesting smart replies based on learned knowledge about a user from different interactions with the same user.
    
\end{enumerate}

\section{Suggesting smart replies for multilingual speakers with code switching}

In linguistics, code-switching or language alternation occurs when a speaker alternates between two or more languages, or language varieties, in the context of a single conversation or situation. Code switching happens where people use either same script for a language e.g. English - Hinglish/ English- Spanglish or different script to type both languages English - Hindi/ English - Spanish. Additionally, the challenge arises from the fact that there is large amount of data in monolingual setting i.e. exclusively using 1 language to type the message while there is limited amount of data for code-switching i.e. using 2 language in the same sentence.

For initial work we start with first generating a large corpus of m-r (message-replies) pair in code-switch format. Then we adapt the Smart replies pipeline to support multiple languages by change in architecture and adding few auxillary task to the bi-encoder approach.

\subsection{Code-Switch Data}

The code switch data is constructed by first taking the English m-r (message-replies) pair data. Then each m-r is translated in the monolingual second language e.g. Hindi using an existing solution like Google Translate. Then we take an input English message break it into subordinate clauses and out of all subordinate clauses we apply clause substitution to Hindi based on estimated frequency of code-switch. Hence we generate code-switch samples in English-Hindi-CS (code switch) format.

While the m-r pair data is usually retrieved from user collected data in a commercial application, we use the 
public topical chat data where different turkers discuss on an open ended conversation with varying background knowledge provided via wikipedia and reddit article. The data consists of 8628 conversations and over 184,000 messages across 7 sentiments — Angry, Curious to Dive Deeper, Disguised, Fearful, Happy, Sad, and Surprised.

\subsection{Code-switch Smart Replies model}

To adapt the smart replies model to multiple language we first start with a multilingual BERT as m-r encoders. Then apart from the normal cosine similarlity between the m-r pair embedding, we add additional auxiliary tasks by adding a translation head and task using the learned embedding to make the model work better. We then train the model on the all the 3 tasks.

\begin{figure}[h]
\centering
\includegraphics[width=1.0\linewidth]{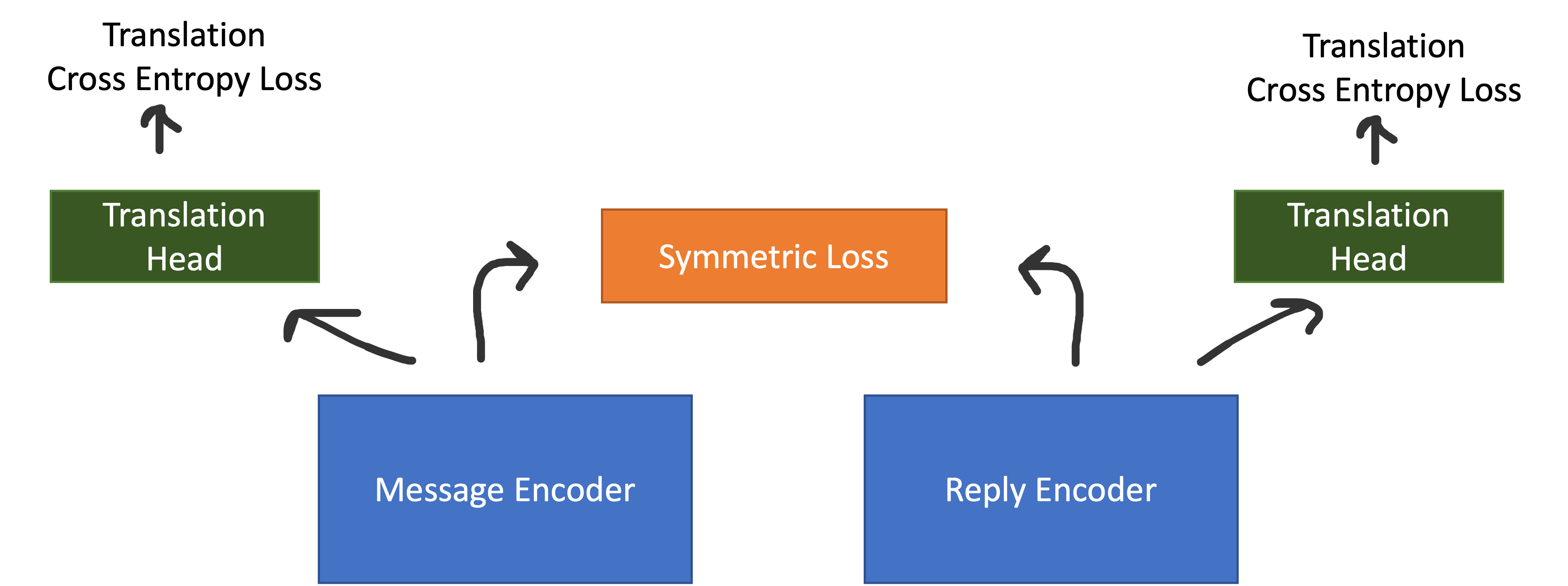}
 \caption{Additional translation task auxillary task to fine-tune message and reply encoder.
}
 \label{fig:smartgmail2}
\end{figure}

\subsection{Results}

To evaluate the results of our Code-switch smart replies. We evaluate the rank of each reply in the response set $R$, based on the score function given in equation \ref{inference}.

Then we sort the set each reply in the
in descending order based on the score. Finally, we find the rank of the actual response with respect to
all elements in $R$.

Using this value, we can compute the Mean Reciprocal Rank:

\[ MRR = \frac{1}{N} \sum_{i=1}^{N} \frac{1}{rank_i} \]

\begin{table}
\centering
\begin{tabular}{|c|c|c|}
  \hline
  Model & MRR & Latency \\
  \hline
  Random & $5.12e-3$ & 15.2 ms \\
  \hline
  English BERT + Translation & $0.431$ & 721.8 ms \\
  \hline
  mBERT on Code-Switch Data & $0.526$ & 340.2 ms \\
  \hline
\end{tabular}
\caption{Results of code-switch ranking}\label{tab:smart_cs_rank} 
\end{table}

Analysing the results in table \ref{tab:smart_cs_rank} we find that the multilingual BERT trained on Code-Switch m-r data performed the best in retrieving the correct response. While using BERT trained on English m-r pair data performed a little worse when we used translation over top of it. This shows that using language based training helps better response generation.

\subsection{Future Directions}

We constructed Code-switch data for training using simple clause based substitution while there exists better methods like embedded matrix theory (EMT) for code-switching data generation which could have been used to construct more natural Code-switch data.

Further we still need to a do a much larger experiments on commercial level data to validate our hypothesis of using multilingual representations over language specific representations.

Another future direction to explore in smart reply space is about Suggesting smart replies based on learned knowledge about a user during the conversation history.

While there has been a lot of research \citep{peng2022godel, zhang2018personalizing} in open dialogue chat domain for generating text based on input text and given context, there is little work on how to model this for human conversation where the external knowledge comes from various interactions between the same 2 users. Usually the external knowledge in existing research is given as input while in our case the external knowledge is actually learned from conversation itself.

E.g. if during a conversation between Alice and Bob, Bob learns that Alice loves pizza, and later on the conversation Alice asks for resturant recommendation, Bob could recommend pizza resturants.

Usually in human interactions we tend to learn more about the other person during the course of multiple interactions. And modify our interactions accordingly based on the learned information. Can we do the same in a smart reply system is another good direction to approach.

\section{Conclusion}

With recent advancements in large language model, our work tries to find applications for the latest research in real world usage. We propose 3 novel on-device tasks to assist users which much more powerful experiences  -  visual question answering, automated form filling using previous on screen context and support for smart replies with linguistic code-switching. We then do initial experiments to propose solutions for each of the task. While the solutions do work on the limited amount of data we have - we observe 2 limitations, lack of large scale experimentation on commercial level, large amount of latency for proposed solutions. Lastly, we also discuss few possible directions for exploration of future work in the field.


\bibliography{acl2020}
\bibliographystyle{acl_natbib}

\end{document}